\documentclass[journal]{IEEEtran}
\usepackage{amssymb}

\usepackage{pifont,calc}
\usepackage{cite}

\usepackage{graphicx}
\ifCLASSINFOpdf
   \usepackage[pdftex]{graphicx}
\else
 \fi
\usepackage[cmex10]{amsmath}
\usepackage{array}

\usepackage{cases} 
\usepackage{bm} 
\usepackage{subfigure}
\usepackage{mathrsfs}
\usepackage{url}
\usepackage{algorithm}
\usepackage{algorithmic}
\usepackage{multirow}
\makeatletter

\newcommand{\Rmnum}[1]{\expandafter\@slowromancap\romannumeral #1@}
\makeatother

\ifCLASSINFOpdf
\else
\fi
\begin{document}
%
\title{Efficient Inverse-Free Incremental and Decremental Algorithms for Multiple Hidden Nodes  in  Extreme Learning Machine}
%
%
%

\author{Hufei~Zhu
\thanks{H. Zhu is with the College of Computer Science and Software, Shenzhen University, Shenzhen 518060, China (e-mail:
zhuhufei@szu.edu.cn).}
}

%
%

\markboth{Journal of \LaTeX\ Class Files,~Vol.~14, No.~8, August~2015}%
{Shell \MakeLowercase{\textit{et al.}}: Bare Demo of IEEEtran.cls for IEEE Journals}

\maketitle

\begin{abstract}
The inverse-free extreme learning machine (ELM) algorithm proposed in \cite{Existing_Inverse_free_best_ELM}
was based on an inverse-free algorithm to compute the regularized pseudo-inverse,  which was deduced
from an inverse-free recursive algorithm to update the inverse of a Hermitian matrix.
Before that recursive algorithm was applied in \cite{Existing_Inverse_free_best_ELM},
its improved version had been utilized  in previous literatures~\cite{zhfICC2009, TransComm2010ReCursiveGstbc}.
Accordingly from the improved recursive algorithm \cite{zhfICC2009, TransComm2010ReCursiveGstbc},
several  efficient  inverse-free algorithms for ELM were proposed in \cite{my_submitted_ELM1}
to  reduce the computational complexity. In this paper, we propose two  inverse-free algorithms  for ELM with Tikhonov
regularization, which can increase multiple hidden nodes in an iteration.  On the other hand,   we also propose two  efficient decremental learning algorithms for ELM with Tikhonov
regularization, which can remove multiple redundant nodes in an iteration.
\end{abstract}


%
\IEEEpeerreviewmaketitle

\section{Introduction}

The extreme learning machine (ELM)~\cite{ELM2004cof}
is an
effective solution
 for Single-hidden-layer feedforward networks (SLFNs)
 due to its unique characteristics,
i.e., extremely fast learning speed, good generalization performance, and
universal approximation capability~\cite{ref17ElmInvFreePaper}.
Thus ELM
has
been
widely applied in classification and
regression~\cite{ELM2012RegressionAndClassify}.

The incremental ELM proposed in \cite{ref17ElmInvFreePaper}
achieves the universal approximation capability by
adding hidden nodes one by one.
However, it only updates the output weight  for
the newly added hidden node, and freezes the output weights of the existing hidden nodes.
Accordingly  those output weights
 are no longer the optimal least-squares solution of the standard ELM algorithm.
 Then the inverse-free algorithm was proposed in \cite{Existing_Inverse_free_best_ELM}
to  update the output weights of  the added node and the existing nodes simultaneously,
 and the updated weights are identical to the optimal solution of the standard ELM algorithm.
 The ELM algorithm in \cite{Existing_Inverse_free_best_ELM} was based on
an inverse-free algorithm to compute the regularized pseudo-inverse,  which was deduced
from an inverse-free recursive algorithm to update the inverse of a Hermitian matrix.

Before the recursive algorithm to update the inverse was utilized in \cite{Existing_Inverse_free_best_ELM},
it had been mentioned in previous literatures~\cite{Matrix_Handbook,Matrix_SP_book,GlobecommRecursiveVBLAST, TransWC2009ReCursiveBlast,zhfICC2009},
while its improved version had been utilized  in \cite{zhfICC2009, TransComm2010ReCursiveGstbc}.
Accordingly from the improved recursive algorithm  \cite{zhfICC2009, TransComm2010ReCursiveGstbc},
several  efficient  inverse-free algorithms for ELM were proposed in \cite{my_submitted_ELM1}
to  reduce the computational complexity.
In \cite{my_submitted_ELM1},  two proposed  inverse-free ELM algorithms compute the output weights directly from the
 inverse and the ${\bf{LD}}{{\bf{L}}^T}$
   factors of the inverse, respectively, to
 avoid computing the regularized pseudo-inverse that is usually bigger than the
 inverse.

In each iteration, the inverse-free algorithms in  \cite{Existing_Inverse_free_best_ELM, my_submitted_ELM1} for ELM with Tikhonov
regularization can only increase one hidden node. In this paper, we develop two  inverse-free algorithms  for ELM with Tikhonov
regularization, which can increase multiple hidden nodes in an iteration.  On the other hand,  it is  also required to prune redundant nodes~\cite{NodeDecremental1,NodeDecremental2,NodeDecremental3,NodeDecremental4,NodeDecremental5,NodeDecremental6}
 by decremental learning algorithms in machine learning.
 Thus in this paper,  we will propose  efficient decremental learning algorithms to remove multiple redundant nodes in an iteration.

\section{Architecture of the ELM}

In the ELM model, the
 $n$-th input node, the $i$-th hidden node, and the $m$-th output node can be denoted as $x_n$, ${h_i}$, and ${z_m}$, respectively,
 while all the $N$ input nodes, $l$ hidden nodes, and $M$ output nodes
 can be denoted as ${\bf{x}} = {\left[ {\begin{array}{*{20}{c}}
{{x_{1}}}&{{x_{2}}}& \cdots &{{x_{N}}}
\end{array}} \right]^T} \in {\Re ^N}$,
 ${\bf{h}} = {\left[ {\begin{array}{*{20}{c}}
{{h_{1}}}&{{h_{2}}}& \cdots &{{h_l}}
\end{array}} \right]^T} \in {\Re ^l}$,
and ${\bf{z}} = {\left[ {\begin{array}{*{20}{c}}
{{z_{1}}}&{{z_{2}}}& \cdots &{{z_{M}}}
\end{array}} \right]^T} \in {\Re ^M}$, respectively.
Accordingly
 the ELM model  can be represented  in a compact form
as
 \begin{equation}\label{equ03}
{\bf{h}} = f\left( {{\bf{Ax}} + {\bf{d}}} \right)
 \end{equation}
 and
  \begin{equation}\label{equ04}
{\bf{z}} = {\bf{Wh}},
 \end{equation}
 where
${\bf{A}} = \left[ {{a_{in}}} \right] \in {\Re ^{l \times N}}$,
${\bf{d}} = {\left[ {\begin{array}{*{20}{c}}
{{d_1}}&{{d_2}}& \cdots &{{d_l}}
\end{array}} \right]^T} \in {\Re ^l}$,
${\bf{W}} = \left[ {{w_{mi}}} \right] \in {\Re ^{M \times l}}$,
and  the activation
function $f( \bullet )$ is
entry-wise, i.e., $f({\bf{A}}) = \left[ {f({a_{in}})} \right] \in {\Re ^{l \times N}}$ for a matrix input ${\bf{A}} = \left[ {{a_{in}}} \right] \in {\Re ^{l \times N}}$.
In (\ref{equ03}),  the activation
function $f( \bullet )$
can be chosen as linear, sigmoid, Gaussian
models, etc.

Assume there are totally $K$ distinct   training samples, and let ${{\bf{x}}_{k}}\in {\Re ^N}$ and ${{\bf{z}}_{k}}\in {\Re ^M}$ denote the $k$-th training input and the corresponding $k$-th training output, respectively, where $k=1,2,\cdots,K$. Then the input sequence and the output sequence in the training set can be represented as
\begin{equation}\label{equ05}
{\bf{X}} = \left[ {\begin{array}{*{20}{c}}
{{{\bf{x}}_{1}}}&{{{\bf{x}}_{2}}}& \cdots &{{{\bf{x}}_{K}}}
\end{array}} \right] \in {\Re ^{N \times K}},
\end{equation}
and
 \begin{equation}\label{equ06}
{\bf{Z}} = \left[ {\begin{array}{*{20}{c}}
{{{\bf{z}}_{1}}}&{{{\bf{z}}_{2}}}& \cdots &{{{\bf{z}}_{K}}}
\end{array}} \right] \in {\Re ^{M \times K}},
 \end{equation}
respectively. We can substitute (\ref{equ05}) into (\ref{equ03}) to obtain
 \begin{equation}\label{equ07}
{\bf{H}} = f\left( {{\bf{AX}} + {{\bf{1}}^T} \otimes {\bf{d}}} \right),
  \end{equation}
where ${\bf{H}} = \left[ {\begin{array}{*{20}{c}}
{{{\bf{h}}_{1}}}&{{{\bf{h}}_{2}}}& \cdots &{{{\bf{h}}_{K}}}
\end{array}} \right] \in {\Re ^{l \times K}}$ is the value sequence of all $l$ hidden nodes,
and $\otimes$ is the Kronecker product~\cite{Existing_Inverse_free_best_ELM}. Then we can substitute (\ref{equ07}) and (\ref{equ06})
into  (\ref{equ04}) to obtain the actual training
output sequence
\begin{equation}\label{equ08}
{\bf{Z}} = {\bf{WH}}.
\end{equation}

In an ELM, only the output weight ${\bf{W}}$ is adjustable, while ${\bf{A}}$ (i.e., the input weights) and ${\bf{d}}$ (i.e., the biases of the hidden nodes) are randomly fixed.
Denote the desired output as ${\bf{Y}}$.
Then an ELM
simply minimizes the estimation error
\begin{equation}\label{Edefine32178a}
{\bf{E}}={\bf{Y}} - {\bf{Z}}={{\bf{Y}} - {\bf{WH}}}
\end{equation}
by finding a least-squares solution ${\bf{W}}$ for the problem
 \begin{equation}\label{equ09}
\mathop {\min }\limits_{\bf{W}} \left\| {\bf{E}} \right\|_F^2=\mathop {\min }\limits_{\bf{W}} \left\| {{\bf{Y}} - {\bf{WH}}} \right\|_F^2,
 \end{equation}
where
$\left\| {\bullet} \right\|_F$  denotes the Frobenius norm.

For the problem (\ref{equ09}), the unique minimum norm least-squares solution is~\cite{ELM2004cof}
 \begin{equation}\label{equ10}
{\bf{W}} = {\bf{Y}}{{\bf{H}}^T}{\left( {{\bf{H}}{{\bf{H}}^T}} \right)^{ - 1}}.
 \end{equation}
To avoid over-fitting, the popular Tikhonov regularization~\cite{48_regularized,49_regularized} can
be utilized to  modify (\ref{equ10}) into
 \begin{equation}\label{equ11}
{\bf{W}} = {\bf{Y}}{{\bf{H}}^T}{\left( {{\bf{H}}{{\bf{H}}^T}{+}k_0^2{\bf{I}}} \right)^{ - 1}},
 \end{equation}
 where $k_0^2 >0$ denotes the regularization factor.
 Obviously  (\ref{equ10}) is just the special case of (\ref{equ11}) with $k_0^2=0$. Thus in what follows,
 we only consider (\ref{equ11}) for the ELM with Tikhonov regularization.

\section{The Existing Inverse-Free ELM Algorithms in \cite{my_submitted_ELM1}}

In machine learning, it is a common strategy  to
 increase the hidden node number gradually until the desired
accuracy is achieved. However, when this strategy is applied in ELM directly, the matrix inverse operation in (\ref{equ11})
 for the conventional ELM will be required when a few or only one extra hidden node is introduced, and accordingly the algorithm will be computational prohibitive.
Accordingly  an inverse-free strategy was proposed in \cite{Existing_Inverse_free_best_ELM},
to  update the output weights incrementally with the increase of the hidden nodes. In each step,
 the output weights obtained by the inverse-free algorithm
 are identical to the solution of the standard ELM algorithm using the inverse operation.

 Assume that in the ELM with $l$ hidden nodes, we
 add
 one extra hidden node, i.e., the hidden node $l+1$, which has the input weight row vector ${{\bf{\bar a}}_{l + 1}^T} = {\left[ {\begin{array}{*{20}{c}}
{{a_{(l + 1)1}}}&{{a_{(l + 1)2}}}& \cdots &{{a_{(l + 1)N}}}
\end{array}} \right]}$ $\in ({\Re ^N})^T$ and the bias ${\bar d}_{l+1}$.  Then from (\ref{equ07}) it can be seen that
the extra row ${\bf{\bar h}}_{l + 1}^T = f\left( {{\bf{\bar a}}_{l + 1}^T{\bf{X}} + {\bar d}_{l+1}{{\bf{1}}^T} } \right)$ needs to be added to
${\bf{H}} $, i.e.,
 \begin{equation}\label{Haddrow2134}
{{\bf{H}}^{l + 1}} = \left[ \begin{array}{l}
{{\bf{H}}^l}\\
{\bf{\bar h}}_{l + 1}^T
\end{array} \right],
 \end{equation}
 where  ${{\bf{H}}^{i}}$ ($i=l,l+1$)  denotes ${\bf{H}}$ for the ELM with $i$ hidden nodes.
In  ${{\bf{\bar a}}_{l + 1}}$, ${\bf{\bar h}}_{l + 1}$, ${\bar d}_{l+1}$ and what follows, we add the overline
to emphasize the extra vector or scalar, which is added to the matrix or vector for the ELM with $l$ hidden nodes.

After ${\bf{H}} $ is updated by (\ref{Haddrow2134}), the conventional ELM updates the output weights by
  (\ref{equ11}) that involves an inverse
operation.
 To avoid that inverse
operation, the algorithm in \cite{Existing_Inverse_free_best_ELM}
utilizes an inverse-free algorithm to update
 \begin{equation}\label{equ0defB3789}
{\bf{B}} = {{\bf{H}}^T}{\left( {{\bf{H}}{{\bf{H}}^T}{+}k_0^2{\bf{I}}} \right)^{ - 1}}
 \end{equation}
 that is the regularized pseudo-inverse of
${{\bf{H}}}$,
and then substitutes (\ref{equ0defB3789}) into (\ref{equ11}) to
 compute the output weights by
 \begin{equation}\label{equ0DefWbyB32797}
{\bf{W}} = {\bf{Y}}{\bf{B}}.
 \end{equation}
 In \cite{Existing_Inverse_free_best_ELM},
${{\bf{B}}^{l + 1}}$ (i.e.,
 ${\bf{B}}$ for the ELM with $l+1$ hidden nodes) is computed from ${{\bf{B}}^l}$ iteratively.

Let
 \begin{equation}\label{equ0Rdef43415}
{\bf{R}} = {\bf{H}}{{\bf{H}}^T}{+}k_0^2{\bf{I}}
 \end{equation}
and
 \begin{equation}\label{equ0QRrelation3123}
{\bf{Q}} =  {{\bf{R}}^{ - 1}}= {\left( {{\bf{H}}{{\bf{H}}^T}{+}k_0^2{\bf{I}}} \right)^{ - 1}}.
 \end{equation}
Then we can write (\ref{equ0defB3789}) as
 \begin{equation}\label{equ0defB3789ToQ}
{\bf{B}} = {{\bf{H}}^T}{\bf{Q}}.
 \end{equation}
From (\ref{equ0Rdef43415}) we have ${{\bf{R}}^{l + 1}} = {{\bf{H}}^{l + 1}}{({{\bf{H}}^{l + 1}})^T}{+}k_0^2{{\bf{I}}_{l + 1}}$, into which we substitute (\ref{Haddrow2134}) to obtain
\begin{equation}\label{equ0Riter2nd141}
{{\bf{R}}^{l + 1}} = \left[ {\begin{array}{*{20}{c}}
{{{\bf{R}}^l}}&{{{\bf{p}}_l}}\\
{{\bf{p}}_l^T}&{{{\bf{\bar h}}_{l+1}^T}{\bf{\bar h}}_{l+1} + k_0^2}
\end{array}} \right],
\end{equation}
where ${{\bf{p}}_l}$, a column vector with $l$ entries, satisfies
 \begin{equation}\label{equ0vDefine}
{{\bf{p}}_l} = {{\bf{H}}^l}{\bf{\bar h}}_{l + 1}.
 \end{equation}

The inverse-free recursive algorithm in \cite{my_submitted_ELM1}  computes ${{\bf{Q}}^{l + 1}}=({{\bf{R}}^{l + 1}})^{-1}$
by
 \begin{equation}\label{equ0Qgrow3141}
{{\bf{Q}}^{l + 1}} = \left[ {\begin{array}{*{20}{c}}
{{{{\bf{\tilde Q}}}^l}}&{{{\bf{t}}_l}}\\
{{\bf{t}}_l^T}&{{\tau _l}}
\end{array}} \right],
 \end{equation}
where
\begin{subnumcases}{\label{equ0QplusEntry31}}
{{\tau _l}}  = 1/\left( {({{{\bf{\bar h}}_{l+1}^T}{\bf{\bar h}}_{l+1} + k_0^2})  - {\bf{p}}_l^T
{{\bf{Q}}^{l}}
{\bf{p}}_l } \right) &  \label{equ0QplusEntry31a}\\
{\bf{t}}_l  =  - {{\tau _l}} {{\bf{Q}}^{l}} {\bf{p}}_l & \label{equ0QplusEntry31b}\\
{{{{\bf{\tilde Q}}}^l}}  = {{\bf{Q}}^{l}}  + ({1/{\tau _l}})
{\bf{t}}_l {\bf{t}}_l^T. & \label{equ0QplusEntry31c}
\end{subnumcases}
and  update the output weight ${\bf{W}}$ by 
 \begin{equation}\label{W9extendDef232}
 {{\bf{W}}^{l + 1}} = \left[ {\begin{array}{*{20}{c}}
{{{{\bf{\tilde W}}}^l}}&{{\bf{\bar w}}_{l+1}}
\end{array}} \right],
  \end{equation}
where
 \begin{subnumcases}{\label{WplusSimplest41132MyB2myQ}}
{{\bf{\tilde W}}^l} = { {{\bf{W}}^{l}}  + ({\bf{\bar w}}_{l + 1}/{{\tau _l}}) {\bf{t}}_l^T},    &  \label{W2WwpQ321434best}\\
{\bf{\bar w}}_{l+1} = {\tau _l}\left( {\bf{Y}} {\bf{\bar h}}_{l + 1}  - {{\bf{W}}^{l}} {{\bf{p}}_l} \right) & \label{WplusSimplest41132bMyB2myQ}
\end{subnumcases}
are computed from ${\bf{t}}_l$ and ${{\tau _l}}$ in ${{\bf{Q}}^{l + 1}}$.

Since the processor units  are limited in  precision,
the  recursive algorithm utilized to update
 ${\bf{Q}}$
may introduce numerical instabilities,
 which
occurs only after a very large number of iterations~\cite{TransSP2003Blast}.
Thus instead of the inverse of ${\bf{R}}$ (i.e., ${\bf{Q}}$), we can also
update
the inverse ${\bf{LD}}{{\bf{L}}^T}$
   factors~\cite{zhfVTC2010DivFree} of ${\bf{R}}$,
   since usually the ${\bf{LD}}{{\bf{L}}^T}$
   factorization is numerically stable~\cite{Matrix_Computations_book}.
   The inverse ${\bf{LD}}{{\bf{L}}^T}$
   factors include the upper-triangular ${{\bf{L}}}$
   and the diagonal ${\bf{D}}$, which
   satisfy
\begin{equation}\label{LDL2QRinv43053}
 {\bf{LD}}{{\bf{L}}^T} = {\bf{Q}} = {{\bf{R}}^{ - 1}}.
  \end{equation}
  From (\ref{LDL2QRinv43053}) we can deduce
 \begin{equation}\label{LDL2QRinv43053Inverse}
 {{\bf{L}}^{-T}}{{\bf{D}}^{-1}}{{\bf{L}}^{-1}}  = {{\bf{R}}},
  \end{equation}
  where the lower-triangular ${{\bf{L}}^{-T}}$ is the conventional ${\bf{LD}}{{\bf{L}}^T}$ factor~\cite{Matrix_Computations_book} of ${{\bf{R}}}$.

 The inverse ${\bf{LD}}{{\bf{L}}^T}$
   factors can be computed from ${\bf{R}}$ directly by
 the inverse ${\bf{LD}}{{\bf{L}}^T}$ factorization in \cite{zhfVTC2010DivFree},
 i.e.,
 \begin{subnumcases}{\label{LDLdefineLD133}}
{{\bf{L}}^{l + 1}} = \left[ {\begin{array}{*{20}{c}}
{{{\bf{L}}^l}}&{{{{\bf{\tilde t}}}_l}}\\
{{\bf{0}}_l^T}&1
\end{array}} \right] &  \label{LDLdefineLD133L}\\
{{\bf{D}}^{l + 1}} = \left[ {\begin{array}{*{20}{c}}
{{{\bf{D}}^l}}&{{\bf{0}}_l^{}}\\
{{\bf{0}}_l^T}&{{\tau _l}}
\end{array}} \right],  & \label{LDLdefineLD133D}
\end{subnumcases}
where
\begin{subnumcases}{\label{LDLdefineLD133td}}
{{{\bf{\tilde t}}}_l}= - {{\bf{L}}^l} {{\bf{D}}^l} ({{\bf{L}}^l})^T  {\bf{p}}_l &  \label{LDLdefineLD133t}\\
{{\tau _l}}= 1/\left( {({{{\bf{\bar h}}_{l+1}^T}{\bf{\bar h}}_{l+1} + k_0^2})  - {\bf{p}}_l^T
{{\bf{L}}^l} {{\bf{D}}^l} ({{\bf{L}}^l})^T
{\bf{p}}_l } \right). & \label{LDLdefineLD133VECd}
\end{subnumcases}

We can show that
 ${{{\bf{\tilde t}}}_l}$ in (\ref{LDLdefineLD133t}) and ${\bf{t}}_l$  in (\ref{equ0QplusEntry31b})
  satisfy
\begin{equation}\label{tildet2tTau32943}
{{{\bf{\tilde t}}}_l}={\bf{t}}_l/{{\tau _l}},
 \end{equation}
and ${{\tau _l}}$ in (\ref{LDLdefineLD133VECd}) is equal to ${{\tau _l}}$ in (\ref{equ0QplusEntry31a}),
 by substituting (\ref{LDL2QRinv43053}) into (\ref{LDLdefineLD133t}) and (\ref{LDLdefineLD133VECd}), respectively.
After updating ${\bf{L}}$ and ${\bf{D}}$,
we compute the output weight ${\bf{W}}$ by
(\ref{WplusSimplest41132bMyB2myQ}),
 \begin{equation}\label{W2WwpQ321434best4LDL}
{{\bf{\tilde W}}^l} = { {{\bf{W}}^{l}}  + {\bf{\bar w}}_{l + 1} {\bf{\tilde t}}_l^T},
\end{equation}
and
(\ref{W9extendDef232}),
where   (\ref{W2WwpQ321434best4LDL}) is deduced by
  substituting (\ref{tildet2tTau32943}) into (\ref{W2WwpQ321434best}).

\section{The Proposed Inverse-Free ELM Algorithms  to Increase
Multiple Hidden Nodes by One Iteration}

In \cite{TransComm2010ReCursiveGstbc},  ${{\tau _l}}$ was extended to be
the $2 \times 2$ Alamouti~\cite{AlamoutiBlock} sub-block, while ${{{\bf{t}}_l}}$
was extended to be
the block vector consisting of $2 \times 2$ Alamouti sub-blocks.
In this paper, we extend (\ref{Haddrow2134}),
(\ref{equ0Riter2nd141}) and
(\ref{equ0Qgrow3141})
to be
\begin{equation}\label{HblkIncrease1283}
{{\mathbf{H}}^{l+\delta}}=\left[ \begin{matrix}
   {{\mathbf{H}}^{l}}  \\
   {{{\mathbf{\bar{H}}}}^{\delta}}  \\
\end{matrix} \right],
\end{equation}
\begin{equation}\label{Rdef141321}
{{\mathbf{R}}^{l+\delta}}\text{=}\left[ \begin{matrix}
   {{\mathbf{R}}^{l}} & \mathbf{P}  \\
   {{\mathbf{P}}^{T}} & \mathbf{F}  \\
\end{matrix} \right]
\end{equation}
and
\begin{equation}\label{Qdef094f}
{{\mathbf{Q}}^{l+\delta}}=\left[ \begin{matrix}
   {{{\mathbf{\tilde{Q}}}}^{l}} & \mathbf{T}  \\
   {{\mathbf{T}}^{T}} & \mathbf{G}  \\
\end{matrix} \right],
\end{equation}
respectively,
where ${{{\mathbf{\bar{H}}}}^{\delta}} \in {\Re ^{\delta \times K}} $,
$\mathbf{P}\in {\Re ^{l \times \delta}}$, $\mathbf{T} \in {\Re ^{l \times \delta}} $,
 $\mathbf{F}\in {\Re ^{\delta \times \delta}}$, and $\mathbf{G} \in {\Re ^{\delta \times \delta}}$.
 Moreover,  $\mathbf{T}$,  $\mathbf{G}$ and ${{{\mathbf{\tilde{Q}}}}^{l}}$
 in (\ref{Qdef094f}) can be computed by
 \begin{subnumcases}{\label{QblKinvFinal301}}
\mathbf{G}={{\left( \mathbf{F}-{{\mathbf{P}}^{T}}{{\mathbf{Q}}^{l}}\mathbf{P} \right)}^{-1}} &  \label{e4gfpqp}\\
\mathbf{T}=-{{\mathbf{Q}}^{l}}\mathbf{PG}  &  \label{e5tqpg}\\
{{\mathbf{\tilde{Q}}}^{l}}={{\mathbf{Q}}^{l}}-{{\mathbf{Q}}^{l}}\mathbf{P}{{\mathbf{T}}^{t}}, & \label{e6qqqpt}
\end{subnumcases}
which will be deduced in Appendix A.

Accordingly we update the output weight ${\bf{W}}$ by
\begin{equation}\label{WfromQbest183}
{{\bf{W}}^{l + \delta}} = {\left[ {\begin{array}{*{20}{c}}
{{{\left( {{{\bf{W}}^l} + ({\bf{Y}}{{({{{\bf{\bar H}}}^\delta})}^T} - {{\bf{W}}^l}{\bf{P}}){{\bf{T}}^T}} \right)}^T}}\\
{{{\left( {({\bf{Y}}{{({{{\bf{\bar H}}}^\delta})}^T} - {{\bf{W}}^l}{\bf{P}}){\bf{G}}} \right)}^T}}
\end{array}} \right]^T},
\end{equation}
which will be deduced in Appendix B.

We can also utilize the ${\bf{LD}}{{\bf{L}}^T}$ factors, which can be written as
\begin{subnumcases}{\label{}}
 {{\mathbf{L}}^{l+\delta}}=\left[ \begin{matrix}
   {{\mathbf{L}}^{l}} & \mathbf{U}  \\
   {{\mathbf{0}}^{T}} & \mathbf{V}  \\
\end{matrix} \right]&  \label{Ldef10438ad}\\
 {{\mathbf{D}}^{l+\delta}}=\left[ \begin{matrix}
   {{\mathbf{D}}^{l}} & \mathbf{0}  \\
   {{\mathbf{0}}^{T}} & \mathbf{S}  \\
\end{matrix} \right], & \label{Ddef034ask}
\end{subnumcases}
where $\mathbf{U}\in {\Re ^{l \times \delta}}$, $\mathbf{V}\in {\Re ^{\delta \times \delta}}$ and $\mathbf{S}\in {\Re ^{\delta \times \delta}}$ can be computed by
\begin{subnumcases}{\label{UVScomputeE3013}}
 \mathbf{U}=-{{\mathbf{L}}^{l}}{{\mathbf{D}}^{l}}{{({{\mathbf{L}}^{l}})}^{T}}\mathbf{PV}  &  \label{e7uldlpv}\\
 \mathbf{VS}{{\mathbf{V}}^{T}}={{\left( \mathbf{F}-{{\mathbf{P}}^{T}}{{\mathbf{Q}}^{l}}\mathbf{P} \right)}^{-1}}. & \label{e8vsvfpqp}
\end{subnumcases}
Accordingly we can update the output weight ${\bf{W}}$ by
\begin{equation}\label{W2wywpvsuuhwpvsv322}
{{\bf{W}}^{l + \delta}} = {\left[ {\begin{array}{*{20}{c}}
{{{\left( {{{\bf{W}}^l} + ({\bf{Y}}{{({{{\bf{\bar H}}}^\delta})}^T} - {{\bf{W}}^l}{\bf{P}}){\bf{VS}}{{\bf{U}}^T}} \right)}^T}}\\
{{{\left( {({\bf{Y}}{{({{{\bf{\bar H}}}^\delta})}^T} - {{\bf{W}}^l}{\bf{P}}){\bf{VS}}{{\bf{V}}^T}} \right)}^T}}
\end{array}} \right]^T}.
\end{equation}
We will deduce (\ref{UVScomputeE3013}) and (\ref{W2wywpvsuuhwpvsv322}) in Appendix C.
Notice that in (\ref{e8vsvfpqp}), the upper-triangular $\mathbf{V}$ and the diagonal $\mathbf{S}$ are the inverse ${\bf{LD}}{{\bf{L}}^T}$ factors of
$\mathbf{F}-{{\mathbf{P}}^{T}}{{\mathbf{Q}}^{l}}\mathbf{P}$
and can be computed by the inverse ${\bf{LD}}{{\bf{L}}^T}$ factorization in \cite{zhfVTC2010DivFree}, or by invert and transpose the traditional ${\bf{LD}}{{\bf{L}}^T}$ factors of $\mathbf{F}-{{\mathbf{P}}^{T}}{{\mathbf{Q}}^{l}}\mathbf{P}$.

\section{The Proposed Inverse-Free ELM Algorithms  to Remove
Multiple Hidden Nodes}


Assume the $\tau $ nodes corresponding to the rows $i_1,i_2,\cdots,i_{\tau}$ ($i_1<i_2<\cdots<i_{\tau}$) in ${{\mathbf{H}}^{l}}$
needs to be removed. Then let us permute the rows $i_1,i_2,\cdots,i_{\tau}$ in ${{\mathbf{H}}^{l}}$
to be the last $1^{st}, 2^{nd}, \cdots, {\tau}^{th}$ rows, respectively,  and the permuted ${{\mathbf{H}}^{l}}$ can be written as
\begin{equation}\label{HblkDecrease1283}
{{\mathbf{H}}^{l}}=\left[ \begin{matrix}
   {{\mathbf{H}}^{l-\tau}}  \\
   {{{\mathbf{{H}}}}^{\tau}}  \\
\end{matrix} \right],
\end{equation}
 where ${{{\mathbf{{H}}}}^{\tau}}$ includes the $\tau$ rows to  be removed.
 Since (\ref{HblkDecrease1283}) and (\ref{HblkIncrease1283})  have the same form, we can write
 (\ref{Qdef094f}) as
\begin{equation}\label{Qdef094fdecrease}
{{\mathbf{Q}}^{l}}=\left[ \begin{matrix}
   {{{\mathbf{\tilde{Q}}}}^{l-\tau}} & {\mathbf{T}}_{\tau}  \\
   {{\mathbf{T}}_{\tau}^{T}} & {\mathbf{G}}_{\tau}  \\
\end{matrix} \right],
\end{equation}
respectively,
where
 ${\mathbf{T}}_{\tau} \in {\Re ^{(l-\tau) \times \tau}} $
 and ${\mathbf{G}}_{\tau} \in {\Re ^{\tau \times \tau}}$.

From (\ref{e5tqpg}) we can deduce  ${{\mathbf{Q}}^{l}}\mathbf{P}=-\mathbf{T} \mathbf{G}^{-1}$, which is substituted into (\ref{e6qqqpt})
to obtain
${{\mathbf{Q}}^{l}}={{\mathbf{\tilde{Q}}}^{l}}- \mathbf{T} \mathbf{G}^{-1}{{\mathbf{T}}^{T}}$, i.e.,
\begin{equation}\label{DecreaseQcompute3229as}
{{\mathbf{Q}}^{l-\tau}}={{\mathbf{\tilde{Q}}}^{l-\tau}}- {\mathbf{T}}_{\tau} {\mathbf{G}}_{\tau}^{-1}{{\mathbf{T}}_{\tau}^{t}}.
 \end{equation}
 Correspondingly ${{{\mathbf{Q}}}^{l-\tau}}$ for the remaining $l-\tau$ nodes can be computed by (\ref{DecreaseQcompute3229as}).

Substitute (\ref{DecreaseQcompute3229as}) into (\ref{Qdef094fdecrease}), and then substitute (\ref{Qdef094fdecrease}) and
(\ref{HblkDecrease1283})  into (\ref{PaperHaveWdefine3131}) to obtain
${\bf{W}}_{}^l = {\bf{Y}}{\left[ {\begin{array}{*{20}{c}}
{{\bf{H}}_{}^{l - \tau }}\\
{{\bf{H}}_{}^\tau }
\end{array}} \right]^T}\left[ {\begin{array}{*{20}{c}}
{{\bf{Q}}_{}^{l - \tau } + {\bf{T}}_\tau ^{}{\bf{G}}_\tau ^{ - 1}{\bf{T}}_\tau ^T}&{{\bf{T}}_\tau ^{}}\\
{{\bf{T}}_\tau ^T}&{{\bf{G}}_\tau ^{}}
\end{array}} \right]$, i.e.,
\begin{multline}\label{Wtotal32981}
{\bf{W}}_{}^l = \\
\left[ {\begin{array}{*{20}{c}}
\begin{array}{l}
{\bf{W}}_{}^{l - \tau } + \\
\left( \begin{array}{l}
{\bf{Y}}{({\bf{H}}_{}^{l - \tau })^T}{\bf{T}}_\tau ^{}\\
 + {\bf{Y}}{({\bf{H}}_{}^\tau )^T}{\bf{G}}_\tau ^{}
\end{array} \right)\\
 \times {\bf{G}}_\tau ^{ - 1}{\bf{T}}_\tau ^T
\end{array}&\begin{array}{l}
{\bf{Y}}{({\bf{H}}_{}^{l - \tau })^T}{\bf{T}}_\tau ^{}\\
 + {\bf{Y}}{({\bf{H}}_{}^\tau )^T}{\bf{G}}_\tau ^{}
\end{array}
\end{array}} \right].
 \end{multline}
 From (\ref{Wtotal32981}) we can deduce
 \begin{subnumcases}{\label{}}
{\bf{W}}_{:,1:l - \tau }^l = {\bf{W}}_{}^{l - \tau } + \left( \begin{array}{l}
{\bf{Y}}{({\bf{H}}_{}^{l - \tau })^T}{\bf{T}}_\tau ^{}\\
 + {\bf{Y}}{({\bf{H}}_{}^\tau )^T}{\bf{G}}_\tau ^{}
\end{array} \right){\bf{G}}_\tau ^{ - 1}{\bf{T}}_\tau ^T & \label{decreaseWtotal2a}\\
{\bf{W}}_{:,l - \tau  + 1:l}^l = {\bf{Y}}{({\bf{H}}_{}^{l - \tau })^T}{\bf{T}}_\tau ^{} + {\bf{Y}}{({\bf{H}}_{}^\tau )^T}{\bf{G}}_\tau ^{}, & \label{decreaseWtotal2b}
\end{subnumcases}
where ${\bf{W}}_{:,1:l - \tau }^l$ and  ${\bf{W}}_{:,l - \tau  + 1:l}^l$ denote the first $l - \tau$ columns and the last $\tau$ columns
in ${\bf{W}}_{}^l$, respectively.  Then we can substitute
(\ref{decreaseWtotal2b})  into  (\ref{decreaseWtotal2a})  to deduce ${\bf{W}}_{:,1:l - \tau }^l = {\bf{W}}_{}^{l - \tau } + {\bf{W}}_{:,l - \tau  + 1:l}^l{\bf{G}}_\tau ^{ - 1}{\bf{T}}_\tau ^T$, i.e.,
\begin{equation}\label{decreaseWfinal29392}
{\bf{W}}_{}^{l - \tau } = {\bf{W}}_{:,1:l - \tau }^l - {\bf{W}}_{:,l - \tau  + 1:l}^l{\bf{G}}_\tau ^{ - 1}{\bf{T}}_\tau ^T.
\end{equation}

 Permute the rows $i_1,i_2,\cdots,i_{\tau}$ in ${{\mathbf{L}}^{l}}$
to be the last $1^{st}, 2^{nd}, \cdots, {\tau}^{th}$ rows, respectively.
Since the permuted ${{\mathbf{L}}^{l}}$ is no longer triangular, we can utilized the wide-sense Givens rotation described in
Appendix D
to upper-triangularize  ${{\mathbf{L}}^{l}}$, and update ${{\mathbf{D}}^{l}}$ accordingly. Then ${{\mathbf{L}}^{l-\tau}}$
and ${{\mathbf{D}}^{l-\tau}}$  for the remaining $l-\tau$ nodes are
the sub-matrix
 ${{\mathbf{L}}^{l-\tau}}$ in ${{\mathbf{L}}^{l}}$ and
 the sub-matrix ${{\mathbf{D}}^{l-\tau}}$ in ${{\mathbf{D}}^{l}}$, respectively.

\begin{subnumcases}{\label{LDdefDecreaseTotal121}}
 {{\mathbf{L}}^{l}}=\left[ \begin{matrix}
   {{\mathbf{L}}^{l-\tau}} & \mathbf{U}_\tau  \\
   {{\mathbf{0}}^{T}} & \mathbf{V}_\tau  \\
\end{matrix} \right]&  \label{Ldef10438adDecrease}\\
 {{\mathbf{D}}^{l}}=\left[ \begin{matrix}
   {{\mathbf{D}}^{l-\tau}} & \mathbf{0}  \\
   {{\mathbf{0}}^{T}} & \mathbf{S}_\tau  \\
\end{matrix} \right], & \label{Ddef034askDecrease}
\end{subnumcases}
Substitute (\ref{LDdefDecreaseTotal121}) and (\ref{Qdef094fdecrease}) into (\ref{LDL2QRinv43053}) to obtain
 \begin{equation}\label{UVQTGequ239812}
\left[ \begin{matrix}
   \times  & \mathbf{U}_{\tau }^{{}}\mathbf{S}_{\tau }^{{}}\mathbf{V}_{\tau }^{T}  \\
   \times  & \mathbf{V}_{\tau }^{{}}\mathbf{S}_{\tau }^{{}}\mathbf{V}_{\tau }^{T}  \\
\end{matrix} \right]=\left[ \begin{matrix}
   \mathbf{\tilde{Q}}_{{}}^{l-\tau } & \mathbf{T}_{\tau }^{{}}  \\
   \mathbf{T}_{\tau }^{T} & \mathbf{G}_{\tau }^{{}}  \\
\end{matrix} \right],
\end{equation}
where $\times$ denotes the irrelevant entries.
From (\ref{UVQTGequ239812}) we can deduce
\begin{subnumcases}{\label{UVQTGequ239812bbb}}
\mathbf{T}_{\tau }^{{}}=\mathbf{U}_{\tau }^{{}}\mathbf{S}_{\tau }^{{}}\mathbf{V}_{\tau }^{T}  & \label{UVQTGequ239812bbb1}\\
\mathbf{G}_{\tau }^{{}}=\mathbf{V}_{\tau }^{{}}\mathbf{S}_{\tau }^{{}}\mathbf{V}_{\tau }^{T}, & \label{UVQTGequ239812bbb2}
\end{subnumcases}
which are then substituted into  (\ref{decreaseWfinal29392}) to obtain
${\bf{W}}_{}^{l - \tau } = {\bf{W}}_{:,1:l - \tau }^l - {\bf{W}}_{:,l - \tau  + 1:l}^l{({\bf{V}}_\tau ^{}{\bf{S}}_\tau ^{}{\bf{V}}_\tau ^T)^{ - 1}}{({\bf{U}}_\tau ^{}{\bf{S}}_\tau ^{}{\bf{V}}_\tau ^T)^T}$, i.e,
    \begin{equation}\label{wDecreaseFromLDL32dke}
{\bf{W}}_{}^{l - \tau } = {\bf{W}}_{:,1:l - \tau }^l - {\bf{W}}_{:,l - \tau  + 1:l}^l{\bf{V}}_\tau ^{ - T}{\bf{U}}_\tau ^T.
 \end{equation}

The wide-sense Givens rotation  ${\bf{\Psi
}}_k^{j,j + 1}$ is equal to ${\bf{I}}_k$ except the $2\times 2$
sub-block in the $j^{th}$ and $(j+1)^{th}$ rows and columns, which
is
\begin{equation}\label{S18:AftNonDef_psi}
\left[\setlength{\arraycolsep}{0.3mm}
\renewcommand{\arraystretch}{1.0}
{\begin{array}{*{20}c}
   {\bf{\psi }}_{j,j}^{k,j} & {\bf{\psi }}_{j,j+1}^{k,j}  \\
   {\bf{\psi }}_{j+1,j}^{k,j} & {\bf{\psi }}_{j+1,j+1}^{k,j} \\
\end{array}}\right]=
\frac{1}{\rho} \left[{\begin{array}{*{20}c}
      {l_k^{j+1}} {d_{j+1}}  &   {l_k^{j}} {d_j}   \\
  -{l_k^{j}}{d_{j+1}}  & {l_k^{j+1}}{d_{j+1}} \\
\end{array}}\right],
\end{equation}
where $l_k^i$ is the $i^{th}$ entry in the last $k^{th}$ row of
${\bf{L}}_k$, and
   \begin{equation}\label{rhoDefine38sai}
{\rho}=({l_k^{j+1}} {d_{j+1}})^2+ ({l_k^{j}} {d_j})({l_k^{j}}{d_{j+1}}).
 \end{equation}

\begin{equation}\label{S18:AftNonDef_psiOld}
\left[\setlength{\arraycolsep}{0.3mm}
\renewcommand{\arraystretch}{1.0}
{\begin{array}{*{20}c}
   {\bf{\psi }}_{j,j}^{k,j} & {\bf{\psi }}_{j,j+1}^{k,j}  \\
   {\bf{\psi }}_{j+1,j}^{k,j} & {\bf{\psi }}_{j+1,j+1}^{k,j} \\
\end{array}}\right]=
\frac{1}{\rho} \left[\setlength{\arraycolsep}{0.3mm}
\renewcommand{\arraystretch}{1.0}
{\begin{array}{*{20}c}
     \left| {l_k^{j+1}}  \right|^2{d_{j+1}}  & {{d_j}({l_k^j})^* {l_k^{j+1}}}  \\
  -{d_{j+1}} {{l_k^j}({l_k^{j+1}})^*} & \left| {l_k^{j+1}}  \right|^2{d_{j+1}} \\
\end{array}}\right],
\end{equation}
where $l_k^i$ is the $i^{th}$ entry in the last $k^{th}$ row of
${\bf{L}}_k$, and
   \begin{equation}\label{rhoDefine38saiOld}
{\rho}= \left| {l_k^{j+1}}  \right|\sqrt { \left|{l_k^{j+1}}
\right|^2{d_{j+1}^2} + \left|{l_k^j} \right|^2{d_{j}}{d_{j+1}}}.
 \end{equation}

\section{Conclusion}

In this paper, we develop two  inverse-free algorithms  for ELM with Tikhonov
regularization, which can increase multiple hidden nodes in an iteration.  On the other hand,   we also develop two  efficient decremental learning algorithms for ELM with Tikhonov
regularization, which can remove multiple redundant nodes in an iteration.

\appendices

\section{Derivation of  (\ref{QblKinvFinal301})}

 From
(\ref{equ0QRrelation3123}) we can deduce
 ${\bf{R}}{\bf{Q}} = {\bf{I}}$, into which we can substitute (\ref{Rdef141321})
 and
(\ref{Qdef094f})
to obtain
\begin{equation}\label{RPFQTG2I139481}
\left[ \begin{matrix}
   {{\mathbf{R}}^{l}} & \mathbf{P}  \\
   {{\mathbf{P}}^{T}} & \mathbf{F}  \\
\end{matrix} \right]\left[ \begin{matrix}
   {{{\mathbf{\tilde{Q}}}}^{l}} & \mathbf{T}  \\
   {{\mathbf{T}}^{T}} & \mathbf{G}  \\
\end{matrix} \right]=\mathbf{I}.
\end{equation}
 Then from (\ref{RPFQTG2I139481}) we can obtain
\begin{subnumcases}{\label{QblKinvDeduce01831a}}
{{\mathbf{R}}^{l}}\mathbf{T}+\mathbf{PG}=\mathbf{0} &  \label{e1rtpg}\\
{{\mathbf{P}}^{T}}\mathbf{T}+\mathbf{FG}=\mathbf{I}  &  \label{e2ptfg}\\
{{\mathbf{R}}^{l}}{{\mathbf{\tilde{Q}}}^{l}}+\mathbf{P}{{\mathbf{T}}^{T}}=\mathbf{I}. & \label{e3rqpt}
\end{subnumcases}
From  (\ref{e1rtpg}) we can deduce $\mathbf{T}=-{{({{\mathbf{R}}^{l}})}^{-1}}\mathbf{PG}$, i.e., (\ref{e5tqpg}), which can be substituted into (\ref{e2ptfg})  to obtain $-{{\mathbf{P}}^{T}}{{\mathbf{Q}}^{l}}\mathbf{PG}+\mathbf{FG}=\mathbf{I}$, i.e., (\ref{e4gfpqp}).  Moreover,  from (\ref{e3rqpt})  we can deduce
${{\mathbf{\tilde{Q}}}^{l}}+{{({{\mathbf{R}}^{l}})}^{-1}}\mathbf{P}{{\mathbf{T}}^{T}}={{({{\mathbf{R}}^{l}})}^{-1}}$, i.e., (\ref{e6qqqpt}).

\section{Derivation of  (\ref{WfromQbest183})}

Substitute
(\ref{equ0QRrelation3123})
into
(\ref{equ11}) to obtain
\begin{equation}\label{PaperHaveWdefine3131}
{{\mathbf{W}}^{l+i}}=\mathbf{Y}{{({{\mathbf{H}}^{l+i}})}^{T}}{{\mathbf{Q}}^{l+i}},
\end{equation}
into which substitute (\ref{HblkIncrease1283})
and
(\ref{Qdef094f}) to obtain
${{\mathbf{W}}^{l+i}}=\mathbf{Y}{{\left[ \begin{matrix}
   {{\mathbf{H}}^{l}}  \\
   {{{\mathbf{\bar{H}}}}^{i}}  \\
\end{matrix} \right]}^{T}}\left[ \begin{matrix}
   {{{\mathbf{\tilde{Q}}}}^{l}} & \mathbf{T}  \\
   {{\mathbf{T}}^{T}} & \mathbf{G}  \\
\end{matrix} \right]$,  i.e.,
\begin{multline}\label{w2yhqyhtetc4735}
{{\bf{W}}^{l + i}} =  \\
 \left[ {\begin{array}{*{20}{c}}
{\left( \begin{array}{l}
{\bf{Y}}{({{\bf{H}}^l})^T}{{{\bf{\tilde Q}}}^l} + \\
{\bf{Y}}{({{{\bf{\bar H}}}^i})^T}{{\bf{T}}^T}
\end{array} \right)}&{\left( \begin{array}{l}
{\bf{Y}}{({{\bf{H}}^l})^T}{\bf{T}} + \\
{\bf{Y}}{({{{\bf{\bar H}}}^i})^T}{\bf{G}}
\end{array} \right)}
\end{array}} \right].
\end{multline}

Substitute (\ref{e6qqqpt}) into the first entry in the right side of (\ref{w2yhqyhtetc4735})  to write it as
\begin{equation}\label{yhqyhqptyht8423}
\mathbf{Y}{{({{\mathbf{H}}^{l}})}^{T}}{{\mathbf{Q}}^{l}}-\mathbf{Y}{{({{\mathbf{H}}^{l}})}^{T}}{{\mathbf{Q}}^{l}}\mathbf{P}{{\mathbf{T}}^{t}}+\mathbf{Y}{{({{\mathbf{\bar{H}}}^{i}})}^{T}}{{\mathbf{T}}^{T}},
\end{equation}
and  substitute  (\ref{e5tqpg})   into the second entry in the right side of (\ref{w2yhqyhtetc4735})   to write it as
\begin{equation}\label{yhgyhqpg835423}
\mathbf{Y}{{({{\mathbf{\bar{H}}}^{i}})}^{T}}\mathbf{G}-\mathbf{Y}{{({{\mathbf{H}}^{l}})}^{T}}{{\mathbf{Q}}^{l}}\mathbf{PG}.
\end{equation}
Substitute (\ref{PaperHaveWdefine3131}) into (\ref{yhqyhqptyht8423})  and (\ref{yhgyhqpg835423})
to write them as
\begin{equation}\label{wyhwpt9322}
{{\mathbf{W}}^{l}}+\left( \mathbf{Y}{{({{{\mathbf{\bar{H}}}}^{i}})}^{T}}-{{\mathbf{W}}^{l}}\mathbf{P} \right){{\mathbf{T}}^{T}}
\end{equation}
and
\begin{equation}\label{yhwpg833346}
\left( \mathbf{Y}{{({{{\mathbf{\bar{H}}}}^{i}})}^{T}}-{{\mathbf{W}}^{l}}\mathbf{P} \right)\mathbf{G},
\end{equation}
respectively.
Then the first and second entries in the right side of (\ref{w2yhqyhtetc4735})  can be replaced with
(\ref{wyhwpt9322})  and (\ref{yhwpg833346}), respectively, to obtain (\ref{WfromQbest183}).

\section{Derivation of  (\ref{UVScomputeE3013}) and (\ref{W2wywpvsuuhwpvsv322})}

Substitute (\ref{Ldef10438ad}),  (\ref{Ddef034ask}) and  (\ref{Qdef094f})
into (\ref{LDL2QRinv43053})
 to obtain
$\left[ \begin{matrix}
   {{\mathbf{L}}^{l}} & \mathbf{U}  \\
   {{\mathbf{0}}^{T}} & \mathbf{V}  \\
\end{matrix} \right]\left[ \begin{matrix}
   {{\mathbf{D}}^{l}} & \mathbf{0}  \\
   {{\mathbf{0}}^{T}} & \mathbf{S}  \\
\end{matrix} \right]{{\left[ \begin{matrix}
   {{\mathbf{L}}^{l}} & \mathbf{U}  \\
   {{\mathbf{0}}^{T}} & \mathbf{V}  \\
\end{matrix} \right]}^{T}}=\left[ \begin{matrix}
   {{{\mathbf{\tilde{Q}}}}^{l}} & \mathbf{T}  \\
   {{\mathbf{T}}^{T}} & \mathbf{G}  \\
\end{matrix} \right]$, i.e.,
\begin{equation}\label{LDLbig2QbigE10843}
\left[ \begin{matrix}
   {{\mathbf{L}}^{l}}{{\mathbf{D}}^{l}}{{({{\mathbf{L}}^{l}})}^{T}}+\mathbf{US}{{\mathbf{U}}^{T}} & \mathbf{US}{{\mathbf{V}}^{T}}  \\
   {{(\mathbf{US}{{\mathbf{V}}^{T}})}^{T}} & \mathbf{VS}{{\mathbf{V}}^{T}}  \\
\end{matrix} \right]=\left[ \begin{matrix}
   {{{\mathbf{\tilde{Q}}}}^{l}} & \mathbf{T}  \\
   {{\mathbf{T}}^{T}} & \mathbf{G}  \\
\end{matrix} \right].
\end{equation}
From (\ref{LDLbig2QbigE10843})  we can obtain
\begin{subnumcases}{\label{}}
\mathbf{VS}{{\mathbf{V}}^{T}}=\mathbf{G}  &  \label{e9vsv2g9434}\\
\mathbf{US}{{\mathbf{V}}^{T}}=\mathbf{T}. & \label{e10usv2t05443}
\end{subnumcases}
We can substitute
(\ref{e4gfpqp})
  into (\ref{e9vsv2g9434})   to deduce (\ref{e8vsvfpqp}).  On the other hand, we can substitute   (\ref{e5tqpg})   into (\ref{e10usv2t05443}) to obtain
 $\mathbf{US}{{\mathbf{V}}^{T}}=-{{\mathbf{Q}}^{l}}\mathbf{PG}$, into which we substitute (\ref{e9vsv2g9434})  to obtain
$\mathbf{US}{{\mathbf{V}}^{T}}=-{{\mathbf{Q}}^{l}}\mathbf{PVS}{{\mathbf{V}}^{T}}$, i.e., $\mathbf{U}=-{{\mathbf{Q}}^{l}}\mathbf{PV}$, into which we can substitute (\ref{LDL2QRinv43053})  to obtain
(\ref{e7uldlpv}).

Moreover, 
to deduce (\ref{W2wywpvsuuhwpvsv322}),
 we only need to substitute    (\ref{e9vsv2g9434})    and    (\ref{e10usv2t05443})    into  (\ref{WfromQbest183}).

\section{The Derivation of the Wide-sense Givens Rotation}

In this section we derive the wide-sense Givens rotation ${\bf{\Psi }}_k^{j,j + 1}$
that is
 computed by (\ref{rhoDefine38sai}) and
(\ref{S18:AftNonDef_psi}).
Let ${{\bf{D}}_k^{\frac{1}{2}}}$ denote a diagonal matrix of real
valued weights $\sqrt{d_i}=d_i^{\frac{1}{2}}$. Then we can represent
(\ref{LDL2QRinv43053}) as
\begin{equation}\label{equS17:DeposP2Gam}
{\bf{\bar Q}}_{|k}={{\bf{L}}_{|k} {{\bf{D}}_k^{\frac{1}{2}}}
{{\bf{D}}_k^{\frac{1}{2}}}{\bf{L}}_{|k}^H}.
\end{equation}
Denote the $i^{th}$ column of ${\bf{L}}_{|k}$ as ${\bf{l}}_i$, and
define
\begin{equation}\label{equS17:DefPsqrt}
{\bf{\tilde L}}_k  = {\bf{L}}_{|k} {{\bf{D}}_k^{\frac{1}{2}}}  =
\left[{\begin{array}{*{20}c}
   {{\bf{l}}_1 \sqrt {d_1 } } & {{\bf{l}}_2 \sqrt {d_2 } } &  \cdots  & {{\bf{l}}_k \sqrt {d_k } }  \\
\end{array}} \right].
\end{equation}
Then substitute (\ref{equS17:DefPsqrt}) into
(\ref{equS17:DeposP2Gam}) to obtain ${\bf{\bar Q}}_{|k}  =
{\bf{\tilde L}}_k  {{\bf{\tilde L}}_k^H}$, from which we can deduce
\begin{equation}\label{S18:DefPsqrtUnitTran}
{\bf{\bar Q}}_{|k}  ={\bf{\tilde L}}_k  {{\bf{\tilde L}}_k^H}= {\bf{\tilde L}}_k{\bf{\Omega
}} \left( {{\bf{\tilde L}}_k }{\bf{\Omega}}\right)^H,
\end{equation}
where ${\bf{\Omega }}$ can be any unitary transformation, e.g., a
Givens rotation.


Let ${\tilde l}_k^i$ denotes the $i^{th}$ entry in the last $k^{th}$ row of
${\bf{\tilde L}}_k$. Then let ${\bf{\Omega }}_k^{j,j + 1}$ denote a Givens
rotation that rotates
 $\left[ {\begin{array}{*{20}c}
{{\tilde l}_{k}^{j}  } & {{\tilde l}_{k}^{j+1}}
\end{array}} \right]$ (in the last row of ${\bf{\tilde L}}_k$) into $\left[ {\begin{array}{*{20}c}
0 & {{\tilde {l'}}_{k}^{j+1}}
\end{array}} \right]$. We can employ the efficient Givens rotation
\cite{Complex_Givens_Jul14b}. Then ${\bf{\Omega }}_k^{j,j + 1}$ is
equal to ${\bf{I}}_k$ except the $2\times 2$ sub-block in the
$j^{th}$ and $(j+1)^{th}$ rows and columns, which is
\cite{Complex_Givens_Jul14b}
\begin{equation}\label{equS18:DefeffGiv0}
\frac{1}{\sqrt {\left| {{\tilde l}_k^{j+1}} \right|^2  + \left| {{{\tilde l}_k^j}  }
\right|^2 }} \left[ {\begin{array}{*{20}c}
   \left| {{{\tilde l}_k^{j+1}}} \right| &  {\frac{{{\tilde l}_k^{j+1}}{{{\tilde l}_k^j}}}{\left|
{{{\tilde l}_k^{j+1}} } \right|}}  \\
   { -  {\frac{{{\tilde l}_k^{j+1}}{{{\tilde l}_k^j} }}{\left|
{{{\tilde l}_k^{j+1}} } \right|}}} & \left| {{{\tilde l}_k^{j+1}}} \right|  \\
\end{array}} \right].
\end{equation}

We can obtain ${{\tilde l}_k^i}={l_k^i}\sqrt{d_k}$ from
(\ref{equS17:DefPsqrt}), which is substituted into
(\ref{equS18:DefeffGiv0}) to get
\begin{equation}\label{equS27:DefGivl}
\frac{1}{{\tilde \rho}}\left[ {\begin{array}{*{20}c}
    \left| {l_k^{j+1}}  \right|^2\sqrt{d_{j+1}} & {\sqrt{d_j} {l_k^j} {l_k^{j+1}}} \\
   -{\sqrt {d_j }{l_k^j}{l_k^{j+1}}} &  \left| {l_k^{j+1}}  \right|^2 \sqrt{d_{j+1}}  \\
\end{array}} \right],
\end{equation}
where
   \begin{equation}\label{tildeRhoDefine21918}
{\tilde \rho}=\left| {l_k^{j+1}}  \right|\sqrt { \left|{l_k^{j+1}}
\right|^2{d_{j+1}} + \left|{l_k^j} \right|^2{d_{j}}}.
 \end{equation}

Now we have got ${\bf{\Omega }}_k^{j,j + 1}$. From
(\ref{equS17:DefPsqrt}), we can obtain
\begin{equation}\label{S18:ApplyOme2Psq1}
{\bf{\tilde L}}_k{\bf{\Omega }}_k^{j,j + 1}={\bf{L}}_{|k}
{{\bf{D}}_k^{\frac{1}{2}}}{\bf{\Omega }}_k^{j,j + 1}.
\end{equation}
It can be seen from (\ref{equS27:DefGivl}) that
${{\bf{D}}_k^{\frac{1}{2}}}{\bf{\Omega }}_k^{j,j + 1}$ in (\ref{S18:ApplyOme2Psq1})  is equal to
${{\bf{D}}_k^{\frac{1}{2}}}$ except the $2\times 2$ sub-block in the
$j^{th}$ and $(j+1)^{th}$ rows and columns, which is
\begin{equation}\label{S27:DaftOmega}
\frac{1}{{\tilde \rho}}\left[ {\begin{array}{*{20}c}
    {d_j^{\frac{1}{2}}} \cdot \left| {l_k^{j+1}}  \right|^2{d_{j+1}^{\frac{1}{2}}} & {{d_j^{\frac{1}{2}} }\cdot {d_j^{\frac{1}{2}}}{l_k^j} {l_k^{j+1}}} \\
   -{d_{j+1}^{\frac{1}{2}}} \cdot {{d_j^{\frac{1}{2}} }{l_k^j}{l_k^{j+1}}} &  {d_{j+1}^{\frac{1}{2}}}\cdot \left| {l_k^{j+1}}  \right|^2{d_{j+1}^{\frac{1}{2}}}  \\
\end{array}} \right].
\end{equation}
 Decompose (\ref{S27:DaftOmega}) into
\begin{multline}\label{S27DaftOmegaDec15a}
\frac{1}{{\tilde \rho}} \left[\setlength{\arraycolsep}{0.3mm}
\renewcommand{\arraystretch}{1.0}
{\begin{array}{*{20}c}
     \left| {l_k^{j+1}}  \right|^2{d_{j+1}^{\frac{1}{2}}}  & {{d_j^{\frac{1}{2}} }\cdot {d_j^{\frac{1}{2}}}{l_k^j} {l_k^{j+1}}}/{d_{j+1}^{\frac{1}{2}}}  \\
  -{d_{j+1}^{\frac{1}{2}}} \cdot {{l_k^j}{l_k^{j+1}}} & \left| {l_k^{j+1}}  \right|^2{d_{j+1}^{\frac{1}{2}}} \\
\end{array}}\right] \\
\times \left[\setlength{\arraycolsep}{0.3mm}
\renewcommand{\arraystretch}{1.0}
{\begin{array}{*{20}c}
  {d_j^{\frac{1}{2}}}  &  0 \\
  0  &  {d_{j+1}^{\frac{1}{2}}} \\
\end{array}}\right],
\end{multline}
i.e.,
\begin{multline}\label{S27DaftOmegaDec15b}
\frac{1}{\rho} \left[\setlength{\arraycolsep}{0.3mm}
\renewcommand{\arraystretch}{1.0}
{\begin{array}{*{20}c}
     \left| {l_k^{j+1}}  \right|^2{d_{j+1}^{\frac{1}{2}}}{d_{j+1}^{\frac{1}{2}}}  & {{d_j^{\frac{1}{2}} }\cdot {d_j^{\frac{1}{2}}}{l_k^j} {l_k^{j+1}}}  \\
  -{d_{j+1}^{\frac{1}{2}}} \cdot {{l_k^j}{l_k^{j+1}}}{d_{j+1}^{\frac{1}{2}}} & \left| {l_k^{j+1}}  \right|^2{d_{j+1}^{\frac{1}{2}}}{d_{j+1}^{\frac{1}{2}}} \\
\end{array}}\right] \\
\times \left[\setlength{\arraycolsep}{0.3mm}
\renewcommand{\arraystretch}{1.0}
{\begin{array}{*{20}c}
  {d_j^{\frac{1}{2}}}  &  0 \\
  0  &  {d_{j+1}^{\frac{1}{2}}} \\
\end{array}}\right],
\end{multline}
where
   \begin{equation}\label{rhoCompute1a3298}
{\rho}={{\tilde \rho}}{d_{j+1}^{\frac{1}{2}}}.
 \end{equation}

We can write (\ref{S27DaftOmegaDec15b}) as
\begin{equation}\label{S27DaftOmegaDec15c}
\frac{1}{\rho} \left[\setlength{\arraycolsep}{0.3mm}
\renewcommand{\arraystretch}{1.0}
{\begin{array}{*{20}c}
     \left| {l_k^{j+1}}  \right|^2{d_{j+1}}  & {{d_j}{l_k^j} {l_k^{j+1}}}  \\
  -{d_{j+1}} {{l_k^j}{l_k^{j+1}}} & \left| {l_k^{j+1}}  \right|^2{d_{j+1}} \\
\end{array}}\right] \left[\setlength{\arraycolsep}{0.3mm}
\renewcommand{\arraystretch}{1.0}
{\begin{array}{*{20}c}
  {d_j^{\frac{1}{2}}}  &  0 \\
  0  &  {d_{j+1}^{\frac{1}{2}}} \\
\end{array}}\right],
\end{equation}
from which we can deduce (\ref{S18:AftNonDef_psi}). Moreover, we can
substitute (\ref{tildeRhoDefine21918}) into
(\ref{rhoCompute1a3298}) to obtain ${\rho}={d_{j+1}^{\frac{1}{2}}} \left| {l_k^{j+1}}  \right|\sqrt { \left|{l_k^{j+1}}
\right|^2{d_{j+1}} + \left|{l_k^j} \right|^2{d_{j}}}$, i.e.,
(\ref{rhoDefine38sai}).

\ifCLASSOPTIONcaptionsoff
  \newpage
\fi

\end{document}